# Beyond still images: Temporal features and input variance resilience

Amir Hosein Fadaei, Mohammad-Reza A. Dehaqani

## Abstract

Traditionally, vision models have predominantly relied on spatial features extracted from static images, deviating from the continuous stream of spatiotemporal features processed by the brain in natural vision. While numerous video-understanding models have emerged, incorporating videos into image-understanding models with spatiotemporal features has been limited. Drawing inspiration from natural vision, which exhibits remarkable resilience to input changes, our research focuses on the development of a brain-inspired model for vision understanding trained with videos. Our findings demonstrate that models that train on videos instead of still images and include temporal features become more resilient to various alternations on input media.

## Introduction

Our brains intricately process visual information in the form of videos. The natural visual system employs video footage or a combination of fused images to construct comprehensive representations of the surroundings, facilitating their classification and recognition. This visual data encapsulates both the spatial characteristics of objects and details within our environment, as well as temporal features that depict the relationships among spatial features over time and because of that, it shows resilience to spatial and temporal changes. The natural vision unfolds through two parallel pathways: the dorsal and ventral streams, both originating in the occipital lobe.[1] The occipital lobe processes fundamental visual elements such as angles and basic shapes. Subsequently, the dorsal stream extends towards the parietal lobe, contributing significantly to our comprehension of positionings and motion, and the ventral pathway courses towards the temporal lobe, engaging in semantic analysis and object detection.[2–4] This intricate process underscores the dynamic interplay of spatial and temporal aspects in constructing our visual understanding.[5,6]

The evolution of video understanding algorithms aims to emulate visual cognition more closely in the human brain.[7] The emulation of natural vision stands as a central objective in both artificial intelligence and cognitive neuroscience. Video understanding models derive significant inspiration from the principles inherent in natural vision.[8] It is noteworthy that our innate visual capabilities are honed by exposure to videos, even influencing the processing of data derived from static images. Despite this, a prevalent trend in image understanding models involves training exclusively on static images, neglecting the temporal dimension of data. The shift toward acknowledging and leveraging temporal data holds promising implications for enhancing the depth and accuracy of models in comprehending dynamic visual content.[9,10]

One remarkable quality of natural vision that researchers strive to replicate in models is its resilience against diverse input variations, culminating in the creation of invariant environmental representations. These environmental changes can manifest in various spatial or temporal alterations. Spatial changes encompass modifications in object appearance or shape, such as variations in lighting, brightness, rotation, translations, presence of obstacles, shifts in colors, or any noise or distortion that may impact each frame. Natural vision excels in object detection and comprehending their relevance across different environmental contexts. Notably, it demonstrates a remarkable ability to navigate through spatial changes, showcasing adaptability to alterations in the visual landscape. This adaptability proves crucial in real-world scenarios where environmental conditions are subject to constant flux. Moreover, natural vision's robustness extends to temporal shifts introduced by factors like changes in perspective, camera angles, and shifts in time, which are prevalent in web and edited videos. Despite these variations, natural vision remains resilient, consistently perceiving the relationships between the data presented in different frames and can plan future actions and movements based on it. [11] This resilience underscores the aspiration to instill comparable adaptability and robustness in artificial models, contributing to their effectiveness in navigating dynamic and unpredictable visual environments.

Despite the introduction of various models and significant progress in achieving accuracy within more confined test environments, image understanding models continue to face challenges in the presence of diverse spatiotemporal variances. Our hypothesis suggests that a significant contributing factor to this limitation lies in the extensive training of models with static information, leading to the loss of crucial temporal features and relationships. The removal of temporal features and information detailing how entities evolve can significantly diminish the efficiency of our models in detecting similar inputs that have undergone different alterations before being presented to the model. This temporal myopia, resulting from an overemphasis on static data during training, hinders the model's ability to generalize and adapt to the dynamic nature of real-world scenarios. To address this limitation, it becomes imperative to reevaluate training methodologies, placing a greater emphasis on preserving and leveraging temporal information. Incorporating a more comprehensive representation of spatiotemporal relationships in the training process holds promise for enhancing the adaptability and robustness of image understanding models across a broader spectrum of real-world scenarios.

The development of invariant representations plays a pivotal role in understanding tasks, intricately connected to the concept of generalization in our models. As previously mentioned, natural learning involves the observation and understanding of changes, similarities, and patterns, a process best facilitated through the analysis of videos rather than static images. However, the importance of temporal features, intimately linked to motion and alterations, has not been extensively explored in the training of vision models.

Our work contributes on two fronts. First, we introduce a simple multi-stream model inspired by the natural vision, tailored for image and video understanding, and compare its potential to other similar prominent video understanding models. Second, we explore the model's resilience to common input variations using modified images and videos, drawing comparisons with models trained solely on images. This research demonstrates the role of temporal features in creating invariant representations for image and video understanding.

## Method

In our exploration of the impact of transitioning to spatiotemporal features for creating invariant representations, we implemented a simple multi-stream model based on the existing multi-stream networks. Inspired by the principles of natural vision, the initial introduction of two-stream networks featured separate spatial and temporal channels [12–14]. However, these networks faced limitations in scaling for longer videos. To address this challenge, the Temporal Segment [15] and Deep Temporal Linear Encoding (TLE) [16] networks proposed a segmentation approach, dividing videos into smaller sections to enable aggregation and encoding of the final result for temporal fusion. In earlier iterations, two-stream models utilized intermediary media as inputs for the temporal stream. However, more recent models [17] have introduced novel methods for extracting temporal features. Drawing inspiration from the pathways of natural vision, the Slowfast Network [18] employs two pathways for the extraction of spatiotemporal features. One pathway, known as the Fast stream, provides higher temporal coverage with a high frame rate input, allowing for the extraction of finely tuned temporal-focused spatiotemporal features. The other pathway, referred to as the Slow stream, offers lower temporal coverage with a lower frame rate input, providing a robust foundation for extracting features rich in spatial information. Employing multiple parallel pathways extends beyond spatial and temporal streams of vision. By integrating a mixture of expert classifiers [19], we can incorporate audio information into our network, thus influencing the classification outcomes.

These models feature a dedicated temporal stream alongside a pre-trained image understanding model (spatial stream). The temporal stream is designed to emulate the dorsal pathway observed in natural vision, mirroring the way our brains process temporal information. [20] Concurrently, the pre-existing image understanding channel corresponds to the ventral pathway, focusing on spatial aspects. This dual-stream configuration essentially transforms the model from a single-stream image understanding model to a more robust two-stream video understanding model, aligning seamlessly with the primary objective of this research.

The spatial stream in our implementation comprises a pre-trained ResNet [21,22], renowned for its proficiency in object detection and image classification tasks. In addition, the second stream employs the fusion of multiple sampled frames to generate a more sophisticated representation and a long-term feature repository for the video dataset [23,24]. This feature repository equips the network with the capability to infer relationships between various entities. Within the temporal channel, a fusion layer is implemented, utilizing a slow fusion model [25]. This stream also includes a fully connected layer and a NetVLAD [26] layer to enhance the results following fusion.

We include a third-stream channel to train the network using audio data. The audio stream incorporates fully connected, and NetVLAD layers [26], mirroring the structure of the temporal stream. In this implemented model, the spatial stream is ResNet and the temporal and audio streams are similar to the streams used in the WILLOW model [27]. The outputs from these three distinct streams result in vectors of the same size, denoted as $V_S$, $V_T$, and $V_A$. Employing a Mixture of Expert Classifier in conjunction with context gating, we combine these three vectors using weights $g_S$, $g_T$, and $g_A$. The Mixture of Expert Classifier vector is defined as follows:

$$V = V_S\, g_S + V_T\, g_T + V_A g_A \qquad (1)$$

The weights $g_S$, $g_T$, and $g_A$ are computed dynamically based on the input data. These weights are both trainable and adaptable, contingent upon the input, as determined by the Mixture of Experts Classifier model. They enable the model to flexibly adjust its reliance on each stream according to the contextual information. This adjustment is followed by a context gating mechanism and a classifier layer applied to the output vector generated by combining the results from the various streams.

Context gating functions as an attention-based filter, allowing the network to modulate its confidence in output classes. For instance, in a video featuring skiing, the model may exhibit high confidence in detecting trees. However, if these trees are in the background and hold less relevance, the network can adjust its output confidence accordingly using a self-attention filter. The overarching model schema is visualized in (Fig. 3).

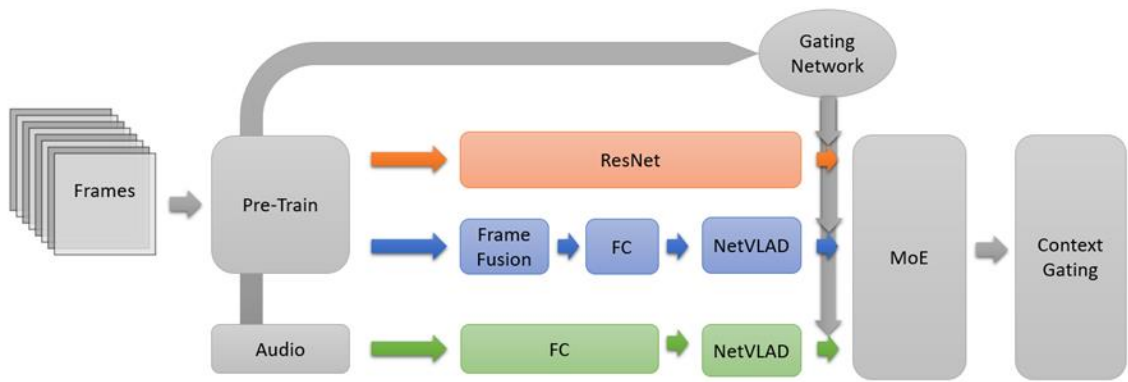

*Figure 3. The model consists of three streams: a spatial stream (depicted in red) takes a single-image input and utilizes a pre-trained ResNet [21,22] for spatial feature detection. The spatiotemporal stream (illustrated in blue) conducts slow fusion on multiple frames to extract spatiotemporal features from the input video. The third stream is an audio stream (depicted in green), contributing significantly to the model's understanding of the input data. The combined output is then processed by the Mixture of Experts classifier and fine-tuned using attention-based context gating.*

## Datasets and Evaluation Metrics

To perform this experiment, various datasets for image and video understanding were employed. For pretraining the image understanding stream and assessing the model's resistance to modified input images, the ImageNet [28] dataset was utilized. Renowned for its extensive coverage, ImageNet [28] is a widely recognized dataset for image understanding and object recognition, comprising over 14 million images across 1000 classes.

To train the model comprehensively on video understanding data, the Youtube-8M [29] dataset was employed, boasting a vast collection of over 6 million videos spanning 3862 classes. Recognized as one of the largest and most extensively validated video understanding datasets, Youtube-8M was chosen as the training set for this model. For assessing the impact of manipulated input videos in video understanding tasks, the Holistic Video Understanding (HVU) [30] dataset, derived from videos in YouTube-8M [29], was utilized. HVU [30] contains over 572000 videos and 3142 classes.

For this study, we generated modified datasets derived from ImageNet [28] and HVU [30] to assess the model's performance in image and video understanding tasks. These modified datasets underwent random spatial alterations, allowing us to evaluate the model's resilience to such changes. Further details on these modified datasets, along with the results of the model's performance on each, are presented in the subsequent section.

To evaluate the model, various metrics, including accuracy, global average precision (GAP) [31], and mean average precision (mAP), were employed. The model generates up to 20 pairs $(e_{v,k}, f_{v,k})$, $k \leq 20$ ; where $e$ represents a class label, and $f$ indicates the confidence level for the predicted class label. Global average precision (GAP) [31] is defined as follows:

$$GAP = \sum_{i=1}^{10000} P(i) \left[ R(i-1) - R(i) \right] \quad (2)$$

Here, $P(i)$ represents the total number of correct labels with confidence above the threshold, divided by the total number of correct classes for all videos at rank $i$. Similarly, $R(i)$ denotes the total number of correct labels with confidence above the threshold, divided by the total number of classes in all videos at rank $i$. The mentioned threshold is dynamically updated based on rank and is calculated as $\tau_i = \frac{i}{10000}$.

mAP@K, which stands for mean average precision ranked at K, is calculated similarly to GAP. It involves computing the average precision for each class, considering the maximum rank of K, and essentially applying the GAP calculation for each class individually.

$$AP_c@K = \sum_{i=1}^{K} P_c(i) \left[ R_c(i-1) - R_c(i) \right] \quad (3)$$

Secondly, we calculate mAP by averaging the resulting average precision (AP) for each class:

$$mAP@K = \frac{1}{C} \sum_{i=1}^{C} AP_c@K \quad (4)$$

Where $C$ represents the total count of classes in the dataset.

# Results

## Training and Evaluation of Three Stream Model

We conducted training on the network using the YouTube-8m dataset [29], where all videos were trimmed to a duration of 10 seconds, equivalent to 300 frames (or less in some cases, depending on the frame ratio of the original data). The spatial network takes the median frame as input, while the temporal network incorporates 30 input frames at a 10:1 sampling ratio. The model was trained with an adaptive learning rate. Our results demonstrate enhanced performance in video understanding tasks in comparison to other models that participated in the YouTube-8m competitions, as presented in (Fig. 4).

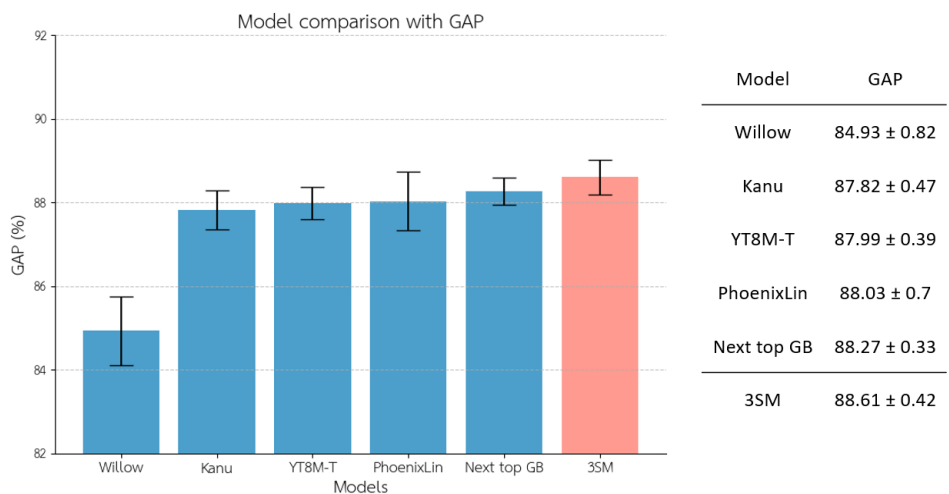

*Figure 4. The performance of the implemented three-stream model (3ST) with audio, spatial, and temporal channels was compared to top-performing models from the YouTube-8M competitions using Global Average Precision (GAP) [31] as the evaluation metric. The objective is to establish the model's relevance, not to assert its status as a state-of-the-art model in video understanding, as that is not the goal of the research.*

## Image Understanding Model

We will utilize two versions of the pre-trained model on Youtube-8M to assess the impact of the temporal stream and evaluate whether the brain-inspired model generates an invariant representation when the temporal stream is included compared to when it is not. In one version, only the spatial stream will be employed, while in the other version, both the spatial and temporal streams will be utilized.

We began by utilizing the ImageNet dataset [28] and subsequently generated modified data from its images. These modifications encompassed several aspects by applying the following probabilities:

   a) 15% probability of reducing image brightness.
   b) 15% probability of rotating the image 45 degrees clockwise.
   c) 15% probability of scaling up the image and selecting the central area across all scales.

d) 55% probability of keeping the image unchanged.

These three modifications address some of the basic spatial alterations. While we trained the two-stream model using videos, our evaluation involved both the ImageNet dataset and the modified ImageNet dataset [28]. For the temporal stream, we used 30 repeated frames of each image as input. Samples of both original and modified images are illustrated in (Fig. 5).

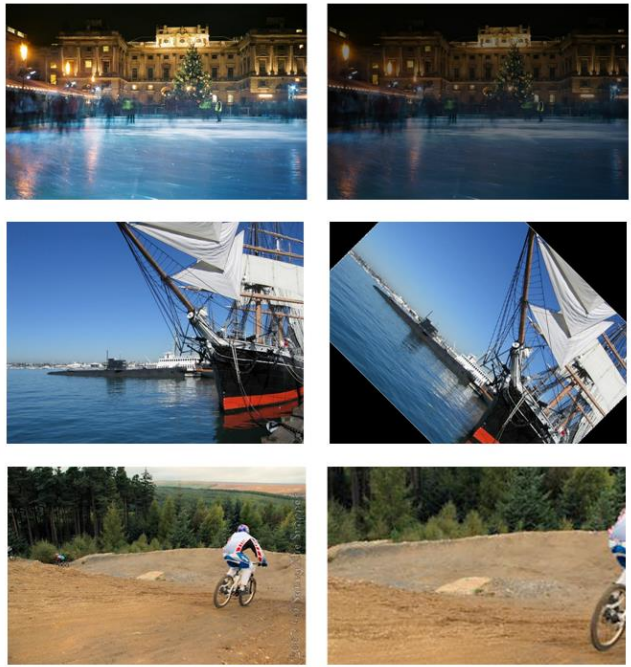

*Figure 5. Here are some examples of modifications applied to the ImageNet dataset, with the modified pictures displayed on the right side. These samples were correctly interpreted by the 2-stream model, whereas the single-stream model struggled to understand them after the modifications.*

This dataset lacks temporal dimension information as it consists of a single repeated image over time. However, this test can demonstrate how a model trained for video understanding, equipped with a dedicated stream for extracting spatiotemporal features, can enhance invariant representations for image understanding tasks. The results indicate that the fine-tuned two-stream model does not surpass the single-stream model in terms of accuracy for image understanding tasks significantly as shown in (Fig. 6). Nevertheless, it is noteworthy that the single-stream model experienced a statistically significant drop in accuracy compared to the two-stream model.

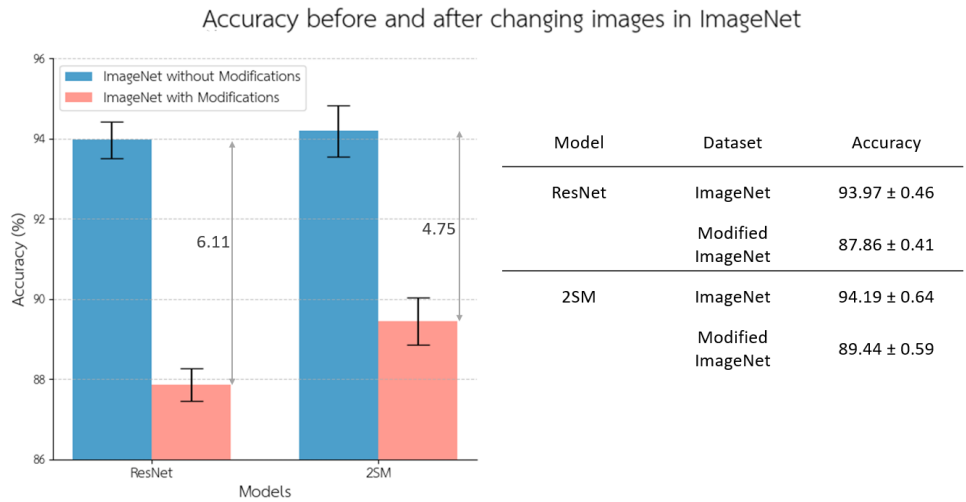

*Figure 6. This plot presents a comparison of accuracy between the two-stream model and the single-stream model (ResNet). We conducted evaluations using an unaltered and modified image understanding dataset (ImageNet), with the modified dataset results being highlighted in red. The findings indicate a notable improvement in accuracy retention when both streams are included, compared to the spatial stream-only model.*

**Video Understanding Model**

Next, we employed the HVU [30] video understanding dataset, which comprises over 9 million training labels, 3142 classes, and more than 572 thousand videos (Some of the videos were not available any more on YouTube at the time of this research). For this test, we introduced modifications to the HVU video dataset, including the following alterations:

a) 10% probability of adding a yellow color filter.
b) 10% probability of adding a red color filter.
c) 10% probability of reducing video brightness.
d) 10% probability of rotating the video 45 degrees counterclockwise.
e) 60% probability of keeping the video unchanged.

Sample frames from the altered video dataset can be viewed in (Fig. 7):

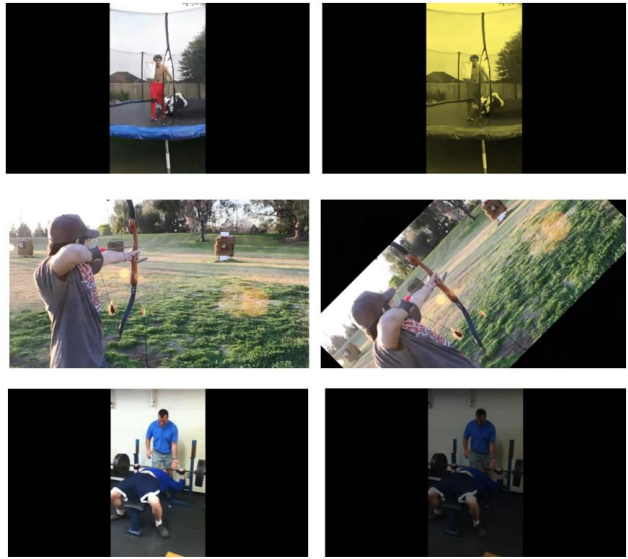

*Figure 7. Here's the HVU dataset along with some examples of the modifications we applied to it, with the modified videos displayed on the right side. These samples were correctly interpreted by the 2-Stream model, whereas the single-stream model struggled to understand them after the modifications.*

We utilized the median frame of the videos for the spatial stream, and we selected 30 frames for the temporal stream. The results indicate an improvement in mAP (mean average precision) [29] when the temporal stream is included, as shown in (Fig. 8). Additionally, there is a notable decrease in mAP when only the spatial stream is used compared to the two-stream model with the modified dataset.

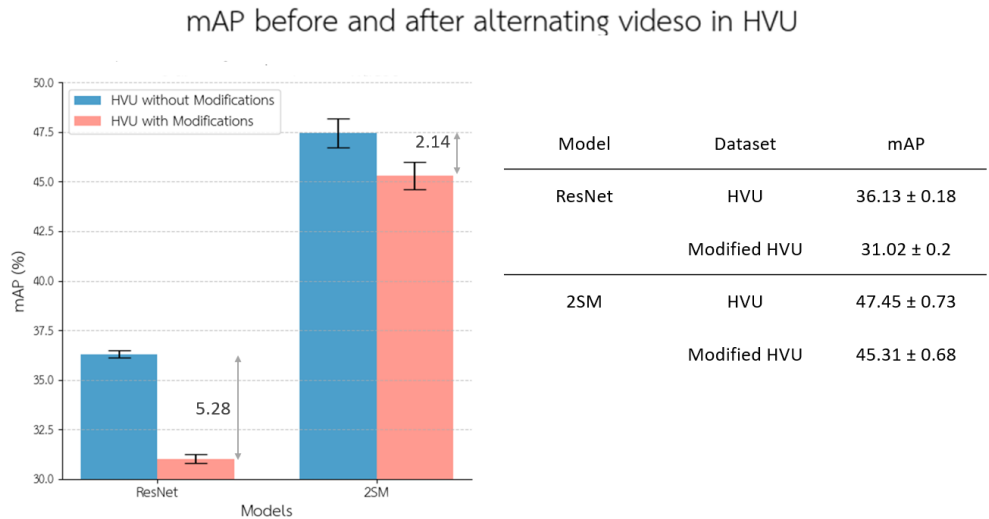

*Figure 8. This table presents a comparison of mAP (mean average precision) between the two-stream model and the single-stream model (ResNet). The evaluations were conducted using both unaltered and modified video understanding datasets, with the modified dataset highlighted in red. The outcomes reveal a significant reduction in the mAP drop when incorporating the spatiotemporal stream, compared to using only the single-stream model.*

In our previous experiments, we employed 30 frames for temporal sampling and fusion out of a total of 300 frames. To further investigate the impact of temporal sampling, we conducted more experiments with varying temporal sampling rates. (Fig. 9) illustrates how the model was influenced by these different temporal sampling rates.

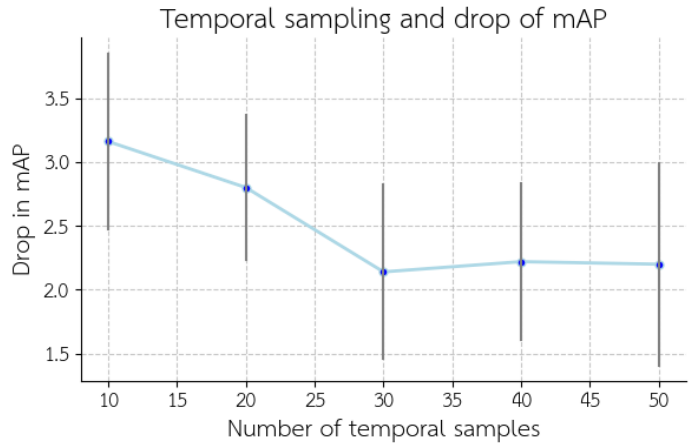

*Figure 9. Increasing temporal sampling does lead to changes in the mAP drop rate. However, it's essential to note that augmenting temporal sampling may not necessarily result in an improved resistance of the model to input variance. The model exhibited the least decline in mAP when using a temporal sampling rate of 30 frames.*

## Discussion

We observe with videos rather than still images. Natural vision exhibits a remarkable tolerance for visual variance. Drawing inspiration from this observation and the results obtained, we can assert that transitioning toward brain-inspired video understanding models with multi-stream spatiotemporal feature extraction will enhance the model's resilience to various input changes. Models focused on spatiotemporal feature extraction and analysis, trained on video datasets, develop a deeper understanding of spatial and temporal variations, thus enabling better generalization for both image and video understanding tasks.

The principal aim of this study is to investigate the challenge of input invariance by contrasting single-stream spatial models with multi-stream spatiotemporal models. In this context, our implementation is exclusively compared to other models that have been previously assessed using the Youtube-8m dataset [29], which allows for such comparisons. Transformer models and even more advanced multi-stream models like the Slowfast Network [18] have the potential to attain superior results in video understanding tasks. However, these structures make it less straightforward to observe and assess the transition from spatial features to spatiotemporal features and their influence on input variance.

Incorporating a second stream for temporal feature extraction leads to the creation of a long-term feature bank, essentially forming a comprehensive impression of the entire video. This 'general impression' proves invaluable in identifying long-term dependencies among spatial features over the temporal domain. Moreover, it enables the model to discern patterns and variations within the spatial domain as they evolve. By including a temporal stream, we enhance the model's robustness against input changes. Much like the dorsal pathway in natural vision, which facilitates understanding of how objects relate to each other and the observer's body in space, the temporal stream within a two-stream model can represent the overarching context of spatial features and their temporal interrelationships.

The results unequivocally demonstrate that training our models with video data enhances their resilience to input variances even in image understanding tasks. Even in cases where the image lacks inherent temporal information, the model can derive substantial benefits from the spatiotemporal structure and the temporal stream. This enables the model to grasp dynamic changes and construct invariant representations, contributing to improved performance in image comprehension.

Multi-stream networks, while not currently at the forefront of video understanding models, have served as valuable tools in our research. Our use of two-stream models allowed us to make meaningful comparisons between image-understanding and video-understanding models for a range of tasks, leveraging their simplicity and structural advantages. Multi-stream models, essentially extensions of image understanding models with temporal fusion, provided an excellent foundation for our research, enabling us to explore the effectiveness of spatiotemporal models in creating invariant representations.

There are several avenues for future research in this field. Exploring the inclusion of an Audio stream and audio data to enhance the creation of invariant representations further. The inclusion of audio and the auditory network is a crucial aspect of natural vision that often complements visual understanding. These sensory inputs work in tandem, enriching the overall perception after undergoing separate processing through parallel pathways. Our model is designed with a three-stream structure, incorporating an audio stream. Drawing inspiration from the principles of natural vision, we hypothesize that audio plays a role in constructing invariant representations of objects and events within the environment. It's worth noting that further dedicated research will be required to delve deeper into this intriguing topic.

Looking ahead, it's crucial to recognize the potential of transformer models with multi-head self-attention, which have attained state-of-the-art results in video understanding.[32] These models have exhibited exceptional performance in achieving generalized understanding and forming invariant representations across various visual tasks. Vision Transformer (ViT)[33] in image understanding and Multiscale Vision Transformers (MViT)[34], TimeSFormer[35], and Video Vision Transformer (ViViT)[36] for Video understanding, are some examples of the transformer-based models for visual understanding. Therefore, one avenue for future research lies in investigating transformer models and their adaptability to input changes, which could shed further light on their capabilities in the context of spatiotemporal modeling.

Additionally, investigating the potential benefits of incorporating feedback mechanisms between spatiotemporal components and layered-wise model comparisons is a subject to consider. Lastly, delving deeper into the alignment between models and the principles of natural vision presents an intriguing area for future investigations.

In summary, our exploration into the realm of spatiotemporal feature extraction has allowed us to cultivate a more generalized understanding and establish invariant representations in our models. This, in turn, equips the model with greater resistance to input variances and changes. To empirically assess this theory, we conducted experiments involving a single-stream spatial model and a two-stream model trained on video data, featuring a dedicated temporal stream. The findings from these experiments validate our hypothesis: the model employing multi-stream spatiotemporal feature extraction exhibits a smaller decline in accuracy when confronted with modified image or video datasets (by more than 3% in mAP for Video understanding tasks). Thus, our results underscore the advantage of training models with video data, even in scenarios where temporal information offers less additional insight for understanding static images.

Our research underscores the importance of designing models and training understanding tasks with videos for achieving better generalization. In essence, our model has demonstrated the ability to create invariant representations applicable to both image and video understanding tasks.

# Data availability

All data supporting the findings of this study are provided within the paper. The original datasets, ImageNet[28] and HVU[30], have been previously published. Details on reconstructing the models used for training and conducting experiments on these datasets, along with information on all random modifications applied to them, are thoroughly described in this article. The modifications were introduced randomly in multiple trials to ensure the statistical significance of the findings.

## Author Information

**School of Electrical and Computer Engineering, College of Engineering, University of Tehran, Tehran, Iran**

Amir Hosein Fadaei, Mohammad-Reza A. Dehaqani

[hosein.fadaei@ut.ac.ir](mailto:hosein.fadaei@ut.ac.ir), [dehaqani@ut.ac.ir](mailto:dehaqani@ut.ac.ir)

## Contributions
Both authors conceived the project and designed the experiments. A.H. Fadaei performed all of the experiments and analyzed the data and wrote the manuscript. M.A. Dehaqani reviewed the manuscript. All authors approved the manuscript for submission.

## Corresponding author

Correspondence to **Mohammad-Reza A. Dehaqani**

## Competing interests

The authors have no relevant financial or non-financial interests to disclose.